# AN IMPROVED GENETIC ALGORITHM WITH A LOCAL OPTIMIZATION STRATEGY AND AN EXTRA MUTATION LEVEL FOR SOLVING TRAVELING SALESMAN PROBLEM

Keivan Borna[1] and Vahid Haji Hashemi[2]

[1]Faculty of Mathematics and Computer Science, Kharazmi University, Tehran, Iran
[2]Faculty of Engineering, Kharazmi University, Tehran, Iran

*ABSTRACT*

*The Traveling salesman problem (TSP) is proved to be NP-complete in most cases. The genetic algorithm (GA) is one of the most useful algorithms for solving this problem. In this paper a conventional GA is compared with an improved hybrid GA in solving TSP. The improved or hybrid GA consist of conventional GA and two local optimization strategies. The first strategy is extracting all sequential groups including four cities of samples and changing the two central cities with each other. The second local optimization strategy is similar to an extra mutation process. In this step with a low probability a sample is selected. In this sample two random cities are defined and the path between these cities is reversed. The computation results show that the proposed method also finds better paths than the conventional GA within an acceptable computation time.*

## 1. INTRODUCTION

The GA maps a set of individual objects or elements, each with a specified value, into a new set of the population [1]. This algorithm attempts to find an approximately good solution to the system by genetically breeding the set of individuals over a series of iterations [2, 3]. The GA algorithm starts by choosing a random set defined as initial population of individuals (a set of solutions) and precedes them in a generational way [4]. During each generation, individuals with high fitness value in the current population are selected to be part of the population formed in the next generation [5, 6]. Generally, the algorithm stops after a fixed number of generations or when an acceptable fitness level has been reached for the last population [7]. The aim of traveling salesman problem (TSP) is to find the shortest tour that passed each city once and exactly once in a known map with different distances between cities. TSP has been widely studied in the fields of artificial intelligence, graph theory, mathematics and computer science due to its applications in real world [14, 15]. However, there are no polynomial algorithms for the NP Complete problems [16]. Exact, approximate and very intelligent methods are extensively designed for TSP. The exact methods waste time and memory and usually are unreachable so local search rules are used to find an approximately good answer. This local search rules are efficient and able to find the shortest or semi shortest path in a polynomial computation time. The local search rules, such as the neighbourhood information [17], may be trap into the local minima and do not find a good answer. Therefore, the quality of the solutions cannot be evaluated due to its random nature and the lack of answer. The intelligent algorithms such as GA are other methods for solving TSP. They find the best or approximate solutions based on the evolutionary rules that differ from the local search rules. In the meantime analytical and some of other intelligent methods can be combined with GA for enhancing its performance. The genetic algorithm is improved by [18] with the reinforcement mutation which relies on the reinforcement learning. The genetic algorithm, firefly method, simulated annealing, ant colony, bee and particle swarm optimization are some of intelligent methods that can be combined with GA for solving TSP [19]. The disadvantage of GA based methods is trap into the local minima. Simple or traditional mutation cannot correct this problem, for this reason GA shortest path is usually has so bigger than the best path. In this paper two local optimization strategies try to improve GA accuracy.

The article is organized as follows. In section 2 GA algorithm phases including crossover and mutation operator studied in TSP. Two local optimization strategies are described with details in Section 3. Section 4 provides an overview of results on a standard TSP dataset. Section 5 highlights the main results of proposed method and indicates further research.

## 2. The genetic algorithm

For solving the TSP with a genetic algorithm, we need a coding, a crossover method, and a mutation method. First of all, algorithm should generate a permutation of integer numbers that each number refers to the $i_{th}$ city in the tour. In this permutation every number may only occur exactly once and belong to interval [1 k], otherwise we do not have a complete tour [8, 9]. The conventional GA one-point crossover method is not inappropriate to do this [10] and some other crossover methods compatible with TSP suggested in [11, 12, 13].

### 2.1. Crossover

#### 2.1.1. Partially Mapped Crossover

Partially mapped crossover (PMX) tries to keep Childs as similar as parents. To achieve this goal, a substring is swapped look like two-point crossover and the values in all other non-conflicting situations are kept. The conflicting positions are changed with the values which swapped to other positions.

An example:

p1 = (1 2 3 4 5 6 7 8 9)    p2 = (4 5 2 1 8 7 6 9 3)

Assume that positions 4–7 are selected for swapping. Then the two offspring's are given as follows if we omit the conflicting positions:

o1 = (* 2 3 | 1 8 7 6 | * 9)   o2 = (* * 2 | 4 5 6 7 | 9 3)

Now we take the conflicting positions and fill in what was swapped to the other offspring. For instance, 1 and 4 were swapped. Therefore, we have to replace the 1 in the first position of o1 by 4, and so on:

o1 = (4 2 3 1 8 7 6 5 9)    o2 = (1 8 2 4 5 6 7 9 3)

**2.1.2. Order Crossover**

Order crossover (OX) is based on this principle that the order of cities is important in compare with its positions in the tour. Similar to PMX, OX swaps two aligned substrings. The computation of the remaining substrings is done with the following way that differs from PMX way. In order to illustrate the OX method, consider the above example (p1, p2) as for PMX. Simply swapping two substrings and omitting all other positions, the result is:

o1 = (* * * | 1 8 7 6 | * *)   o2 = (* * * | 4 5 6 7 | * *)

For computing the open positions of o2, let us write down the positions in p1, but starting from the position after the second crossover site:

9 3 4 5 2 1 8 7 6

If we discard all those values which are already remain in the offspring after swapping (4, 5, 6, and 7), the shortened result is:

9 3 2 1 8

Now OX insert this list into o2 starting after the second crossover position and the updated o2 will be:

o2 = (2 1 8 4 5 6 7 9 3).

Repeating the above process to o1 the following result obtained:
o1 = (3 4 5 1 8 7 6 9 2)

**2.1.3. Cycle Crossover**

PMX and OX usually introduce cities outside the crossover sites which have not been present in either parent. As an example, for instance, the 3 in the first position of o1 in the OX example appears neither in p1 nor in p2. Cycle crossover (CX) tries to overcome this problem and guarantee that every string position in any tour belongs to one of the two parents. Let us continue with the following example:

p1 = (1 2 3 4 5 6 7 8 9)   p2 = (4 1 2 8 7 6 9 3 5)

CX starts from the first position of o1:

o1 = (1 * * * * * * * *)   o2 = (* * * * * * * * *)

o2 may only have a 4 in the first position, because method do not want new values to be introduced there:

o1 = (1 * * * * * * * *)   o2 = (4 * * * * * * * *)

Since the 4 is already fixed for o2 first position, CX keep it in the same position for o1 in order to guarantee that no new positions for the 4 are introduced. We have to keep the 8 in the fourth position of o2 for the same reason:

o1 = (1 * * 4 * * * * *)   o2 = (4 * * 8 * * * * *)

International Journal of Computer Science, Engineering and Information Technology (IJCSEIT), Vol. 4, No.4, August 2014

This process must be repeated for all cities until end up in a value which has previously been considered to complete a cycle:

o1 = (1 2 3 4 * * * 8 *)    o2 = (4 1 2 8 * * * 3 *)

For the second cycle, CX can start with a value from p2 and insert it into o1:

o1 = (1 2 3 4 7 * * 8 *)   o2 = (4 1 2 8 5 * * 3 *)

After the same computations, the result is obtained as following:

o1 = (1 2 3 4 7 * 9 8 5)    o2 = (4 1 2 8 5 * 7 3 9)

The last cycle is a simple replication and the final offspring's are given as follows:

o1 = (1 2 3 4 7 6 9 8 5)    o2 = (4 1 2 8 5 6 7 3 9)

## 2.2 Mutation

Unlike the selection and crossover, in all GA variants similar to real life, the mutation probability is set to a small value [8]. If the mutation probability is set to a large value, the GA is rarely converged and if it is set to a small value, the GA will easily trap into local optima [8]. In this paper, we assign $p_m$ equal to 0.05. In the proposed mutation process a random number assign to each city of a child. If this number is lower than $p_m$, the corresponding city is changed with the second defined city that has a random number lower than $p_m$. If the number of cities lower than $p_m$ in a child was odd, the last of them is discarded. In this process the mutated cities is swapped with each other in a sequential order. This method ensure that no duplicate occur in samples.

## 3. The two local optimization strategy

### 3.1. First local optimization strategy

The first local optimization strategy is extracting all sequential groups including four cities of samples and changing the two central cities with each other. The name of this strategy is four vertices and three lines inequality, which is applied to all samples and the shortest path in each sample, is selected. Based on only two changes in each sample in comparison with main sample, it is not necessary to compute all distances in each process. As a heuristic only the three distances of the selected group should be computed. If this number is lower than initial group, the main sample is reconstruct with this new arrangement and if this number is higher than initial group, algorithm check the next group. For a tour including N cities we have N-3 group with four cities and only three computations is needed to compare sub samples with each other. This means the computational burden of this step is acceptable and can be discarded in total computation time.

### 3.2. Second local optimization strategy

In the ideal situation algorithm should extract and compare all sequential groups including 4 to N-3 cities in each sample and checking all combinations of them with each other.  In a large grid this means a terrible run time. To reduce this complexity, we proposed a new mutation scheme as second local optimization strategy. In the proposed method a random number is assigned to each sample. If this number is lower than $p_m2$, that is selected 0.02 in this paper, the sample was selected for this level of optimization. After selecting a sample, two integer random number between 2 and N-1 are generated (N is the number of cities) and all cities between these numbers will be reversed.

If this new sample tour is lower than initial group, the main sample is reconstructed with this new arrangement and if this number is higher than initial group, algorithm discards it. In simulation process, we can manage run time and accuracy with the $p_m2$ value.

> **The pseudo code of our proposed method**
> Choose initial paths
> Evaluate each path's length
> Determine path's average length
> Repeat
>     Select best-ranking paths to reproduce
>     Mute pairs at random
>     Apply crossover operator
>     Apply level 1 mutation operator
>     Apply level 2 mutation operator
>     Evaluate each path's length
>     Determine path's average lengths
> Until terminating condition
> (E.g. until at least one path has the desired length or enough generations have passed)

## 4. Simulation result

The main parameters of the GA set as follows:

Initial samples are created randomly in search space. The number of initial samples is set to 256. Only N/2 samples with lower tour distance are chosen and other samples are discarded (Ideal selection). We used partially mapped crossover that described in 2.1.1.

The mutation process is similar to 2.2 and $p_m$ is set to 0.05. In the second local strategy the value of $p_m2$ is equal to 0.02. Ftv170 as a standard complicated database is selected from TSPLIB for testing the proposed method and we will compare its efficiency with traditional GA methods. The main parameters for comparing two methods are runtime and answer accuracy. Based on random nature of the total process, we run the two algorithms 30 times and the average results are shows as final results.

Figure 1 shows this result for two algorithms. The tour length in modified method is about 30% lower than its value in conventional GA and obviously shows the improvement in algorithm.
The average runtime of proposed method in 30 times run 1000 iterations, in a Core i5 CPU with 4GB ram is about 140.5 seconds that in compare with 70.4 seconds of conventional GA is about two times higher. Notice that the total run time is negligible and thus the time is not a critical parameter in this comparison.

## 5. Conclusion

In this paper an improved hybrid GA method is used for solving TSP. The proposed method consists of conventional GA and two local optimization strategies. The first local optimization strategy is extracting all sequential groups including four cities of samples and changing the two

51

central cities with each other and is applied to all samples and the shortest path in each sample is selected. The second local optimization strategy is similar to an extra mutation process. In this step with a low probability a sample is selected. In this sample two random cities are defined and the path between these cities is reversed. The computation results show that the proposed method also find the better paths than the conventional GA within an acceptable computation time. In the future we plan to use other meta-heuristic algorithms instead of GA and apply our methods.

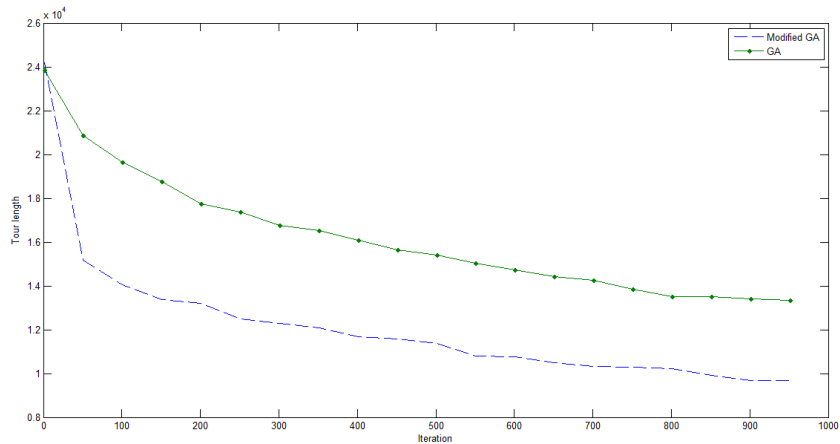

Fig. 1: The tour length vs. iteration number in proposed method and conventional GA

International Journal of Computer Science, Engineering and Information Technology (IJCSEIT), Vol. 4, No.4, August 2014

[16] P. Berman,M, Karpinski, "8/7-Approximation Algorithm for (1,2)-TSP", In: SODA'06, Miami, FL, 2006, pp. 641-648.
[17] Y.H. Liu, "Diversified local search strategy under scatter search framework for the probabilistic traveling salesman problem", European Journal of Operational Research, Vol.191, No.2, 2008, pp. 332-346.
[18] F. Liu, G. Z. Zeng, "Study of genetic algorithm with reinforcement learning to solve the TSP", Expert System with Applications, Vol.36, No.3, 2009, pp. 6995-7001.
[19] S. M. Chen, C. Y. Chien, "Solving the traveling salesman problem based on the genetic simulated annealing ant colony system with particle swarm optimization techniques", Expert System with Applications, Vol.38, No.12, 2011, pp. 14439-14450.
**Authors**

**Dr. Keivan Borna** joined the Department of Computer Science at the Faculty of Mathematics and Computer Science of Kharazmi University as an Assistant Professor in 2008. He earned his Ph.D. in Computational Commutative Algebra from the Department of Mathematics, Statistics and Computer Science of the University of Tehran; where he previously received an M.Sc. in the same field. He also was a visiting scholar in the Dipartemento di Matematicha, Universita' di Genova- Italia and the Department of Mathematik and Informatik at Essen University, Germany, from Sep. 2007 to Apr. 2008. His research interests include Computer Algebra, Cryptography, Approximation Algorithms, and Computational Geometry. He is the author of the "Advanced Programming in JAVA" (in Persian) and is a life member of "Elite National Foundation of Iran". 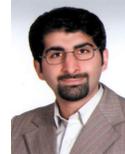

**Vahid Hajihashemi** is currently a master student of Computer Engineering at Faculty of Engineering at Kharazmi University of Tehran. His research interests include artificial intelligence and evolutionary computations. 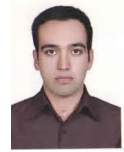

53